\documentclass[11pt]{article}
\usepackage{acl}
\usepackage{times}
\usepackage{latexsym}
\usepackage[T1]{fontenc}
\usepackage[utf8]{inputenc}
\usepackage{microtype}
\usepackage{inconsolata}
\usepackage{graphicx}
\usepackage{booktabs}
\usepackage{multirow}  
\usepackage{amsmath}
\usepackage{makecell}
\usepackage{array}
\usepackage{enumitem}
\usepackage{xcolor}
\usepackage[most]{tcolorbox}

\newif\ifcasecolor
\casecolortrue

\definecolor{EntC}{RGB}{0,92,170}   
\definecolor{RelC}{RGB}{208,120,0}  
\definecolor{PheC}{RGB}{0,128,70}   
\definecolor{StaC}{RGB}{90,90,90}   

\newcommand{\CASEclr}[2]{\ifcasecolor\textcolor{#1}{#2}\else#2\fi}

\newcommand{\ENT}[1]{\textbf{\CASEclr{EntC}{#1}}}
\newcommand{\REL}[1]{\textit{\CASEclr{RelC}{#1}}}
\newcommand{\PHE}[1]{\textbf{\CASEclr{PheC}{#1}}}
\newcommand{\STA}[1]{\CASEclr{StaC}{#1}}

\newcommand{\CaseLegend}{%
\noindent\textbf{Legend.}\;
\parbox[t]{\linewidth}{%
\mbox{\ENT{Entity}\kern0.5em\REL{Relation}\kern0.5em\PHE{Phenotype/Process}}\\[-1pt]
\mbox{\STA{[State/Observation]}.}%
}%
}

\title{BIOME-Bench: A Benchmark for Biomolecular Interaction Inference and Multi-Omics Pathway Mechanism Elucidation from Scientific Literature}

\author{
  \textbf{Sibo Wei}\textsuperscript{1,$\dagger$},
  \textbf{Peng Chen}\textsuperscript{2,$\dagger$},
  \textbf{Lifeng Dong}\textsuperscript{1,3},
  \textbf{Yin Luo}\textsuperscript{1,3},
  \textbf{Lei Wang}\textsuperscript{1,3},
  \textbf{Peng Zhang}\textsuperscript{4},\\
  \textbf{Wenpeng Lu}\textsuperscript{5},
  \textbf{Jianbin Guo}\textsuperscript{1,3,6,$\ast$},
  \textbf{Hongjun Yang}\textsuperscript{2,$\ast$},
  \textbf{Dajun Zeng}\textsuperscript{3,7,$\ast$}
\\
 \textsuperscript{1}Beijing Wenge Technology Co., Ltd, Beijing, China \\
   \textsuperscript{2}Experimental Research Center, China Academy of Chinese Medical Sciences, Beijing, China \\
   \textsuperscript{3}Institute of Automation, Chinese Academy of Sciences, Beijing, China \\
   \textsuperscript{4}College of Intelligence and Computing, Tianjin University, Tianjin, China \\
   \textsuperscript{5}Qilu University of Technology (Shandong Academy of Sciences), Jinan, China \\
   \textsuperscript{6}School of New Media and Communication, Tianjin University, Tianjin, China \\
   \textsuperscript{7}School of Artificial Intelligence, University of Chinese Academy of Sciences, Beijing, China \\
  \small{\textsuperscript{$\dagger$}These authors contributed equally to this work.} \\
  {\small \texttt{*Corresponding authors: \{jianbin.guo, dajun.zeng\}@ia.ac.cn, hjyang@icmm.ac.cn}}
}
\begin{document}
\maketitle
\begin{abstract}
Multi-omics studies often rely on pathway enrichment to interpret heterogeneous molecular changes, but pathway enrichment (PE)-based workflows inherit structural limitations of pathway resources, including curation lag, functional redundancy, and limited sensitivity to molecular states and interventions. Although recent work has explored using large language models (LLMs) to improve PE-based interpretation, the lack of a standardized benchmark for end-to-end multi-omics pathway mechanism elucidation has largely confined evaluation to small, manually curated datasets or ad hoc case studies, hindering reproducible progress. To address this issue, we introduce \textbf{BIOME-Bench}, constructed via a rigorous four-stage workflow, to evaluate two core capabilities of LLMs in multi-omics analysis: \textbf{B}iomolecular \textbf{I}nteraction Inference and end-to-end Multi-\textbf{O}mics Pathway \textbf{M}echanism \textbf{E}lucidation. 
We develop evaluation protocols for both tasks and conduct comprehensive experiments across multiple strong contemporary models. Experimental results demonstrate that existing models still exhibit substantial deficiencies in multi-omics analysis, struggling to reliably distinguish fine-grained biomolecular relation types and to generate faithful, robust pathway-level mechanistic explanations.
Code and datasets are available on GitHub.\footnote{\href{https://github.com/DYJG-research/BIOME-Bench/}{https://github.com/DYJG-research/BIOME-Bench/}}
\end{abstract}

\section{Introduction}
\label{sec:introduction}
Multi-omics profiling, including proteomics, metabolomics, and single-cell transcriptomics, among others, has become central to studying complex biological systems and disease mechanisms ~\cite{sanches2024integrating,baiao2025technical}. By capturing coordinated molecular changes across regulatory layers, multi-omics can support mechanistic hypotheses that connect perturbations to pathway-level phenotypes. However, translating heterogeneous multi-omics signals into coherent, causally grounded explanations remains a major bottleneck ~\cite{mohr2024navigating}.

\begin{figure}
    \centering
    \includegraphics[width=1\linewidth]{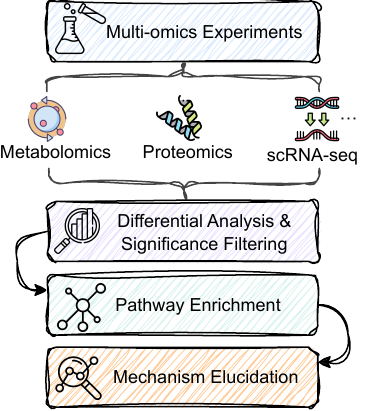}
    \caption{Overview of a pathway enrichment-based multi-omics mechanism elucidation workflow. Multi-omics experiments (e.g., metabolomics, proteomics, and scRNA-seq) are followed by differential analysis with significance filtering to identify perturbed entities, which are then mapped to biological pathways via enrichment analysis to support downstream mechanistic interpretation.}
    \label{fig:case}
\end{figure}

A common approach is pathway enrichment (PE) interpretation ~\cite{zhao2023interpreting,ryan2025pathway,mubeen2022influence}. As shown in Figure~\ref{fig:case}, practitioners typically perform differential analysis and significance filtering to identify perturbed entities, map them to pathways through enrichment, and interpret the enriched pathways to derive mechanistic narratives. PE therefore serves as a widely used interface between entity-level changes and higher-level biological processes ~\cite{elizarraras2024webgestalt}.

However, many limitations of PE-based interpretation arise from the structure of pathway resources. Pathway knowledge bases are curated and versioned, which introduces curation lag and can omit newly discovered or context-specific mechanisms ~\cite{agrawal2024wikipathways}. Enrichment results also exhibit substantial functional redundancy due to overlapping gene sets, often yielding long lists of near-duplicate terms that are difficult to prioritize ~\cite{balestra2023redundancy,ge2025leveraging,ozisik2022orsum}. More fundamentally, enrichment scores are largely context-insensitive. They do not represent molecular states (e.g., phosphorylation or activation), intervention directionality, or the multi-hop causal structure required to connect perturbed entities to pathway-level phenotypes. Consequently, downstream interpretation often relies on post hoc narrative stitching rather than state-aware mechanistic reasoning ~\cite{slobodyanyuk2024directional}. In practice, researchers may obtain an interpretable list of pathway names, but often still need additional analysis to determine how the observed perturbations give rise to the phenotype.

Recent advances in large language models (LLMs) provide opportunities to improve PE-based interpretation, for example by reducing redundancy in enrichment outputs and using LLMs to generate mechanistic explanations~\cite{ge2025leveraging,zhou2024ai}. However, progress in this direction is difficult to measure reliably because existing biomedical benchmarks rarely evaluate the end-to-end capability required to produce mechanistic interpretations from multi-omics observations. Many literature-grounded datasets focus on question answering ~\cite{jin2019pubmedqa} or local relation extraction ~\cite{zhang2019chemical,luo2022sequence}, where relations are often assessed in isolation from pathway context and phenotype-level consequences. Meanwhile, existing pathway-reasoning benchmarks typically emphasize navigation or subgraph operations rather than generating state-aware mechanistic chains that connect perturbed entities to pathway-level phenotypes ~\cite{zhaobenchmarking}. In the absence of such end-to-end benchmarks, evaluations often fall back on small manually curated datasets or ad hoc case studies ~\cite{ge2025leveraging}. This limits coverage and reproducibility, and it leaves open how well current models can directly generate coherent pathway-mechanism explanations on perturbed entities and pathway context.

Motivated by these limitations, we consider a complementary formulation that more closely aligns with how scientists reason from observations to hypotheses: \emph{end-to-end multi-omics pathway mechanism elucidation}. Under this formulation, a model receives only (i) a set of significantly perturbed entities derived from multi-omics measurements and (ii) a pathway context describing the relevant biological process. The model must then generate a coherent mechanistic explanation without relying on explicit pathway retrieval and graph traversal. This end-to-end perspective can reduce redundancy by shifting the unit of interpretation from overlapping pathway labels to mechanistic hypotheses. It also better reflects research practice, where the goal is to elucidate mechanisms from perturbed entities conditioned on pathway context. The task is nontrivial because it requires rich biological knowledge and multi-step causal reasoning to connect perturbed entities to pathway-level phenotypes (see Appendix~\ref{app:case_study}), which motivates standardized, instance-level supervision for reliable evaluation.

To address this gap, we introduce BIOME-Bench, a literature-grounded benchmark designed to evaluate LLMs on two core capabilities: biomolecular interaction inference and end-to-end multi-omics pathway mechanism elucidation. To construct high-quality evaluation instances, we develop a dedicated data construction workflow. As shown in Figure~\ref{fig:workflow}, the workflow transforms pathway information and evidence from biomedical literature into structured, validated knowledge representations through three sequential phases: (i) Literature Search and Relevance Filtering, (ii) Information Extraction and Standardization, and (iii) Knowledge Structuring and Validation. The resulting knowledge representations are then used to formulate the benchmark tasks. Our contributions are as follows:

\begin{itemize}[leftmargin=1.5em]
\item We formulate end-to-end multi-omics pathway mechanism elucidation as a benchmarkable task. Given perturbed entities and pathway context, models must generate coherent, state-aware, and intervention-consistent mechanistic explanations without external retrieval or graph traversal.
\item We construct BIOME-Bench, a literature-grounded benchmark with instance-level supervision that captures key mechanistic elements, including entities, molecular states, biomolecular relations, and pathway-level phenotypes, enabling evaluation beyond surface narrative quality.
\item We design evaluation protocols that measure mechanistic correctness at multiple granularities, including structured knowledge graph evaluation and holistic mechanism evaluation, enabling diagnosis of common failure modes in current LLMs.
\item We benchmark a diverse set of LLMs and analyze their deficiencies, motivating future research toward reliable end-to-end pathway mechanism elucidation from multi-omics observations.
\end{itemize}

\section{Related Work}
\label{sec:related_work}
\subsection{Pathway Knowledge Bases}
KEGG, Reactome, and WikiPathways provide curated pathway representations as molecular interaction and reaction networks~\cite{kanehisa2025kegg, milacic2024reactome, agrawal2024wikipathways}.
KEGG organizes pathway maps using the KO system to support cross-species mapping~\cite{kanehisa2025kegg}. Reactome provides a consistent, manually curated human-centric pathway model with disease and drug context~\cite{milacic2024reactome}. WikiPathways supports community-driven pathway curation~\cite{agrawal2024wikipathways}.
However, these databases are not designed as end-to-end benchmarks for multi-omics mechanism elucidation. They typically lack benchmark-ready, instance-level supervision, such as explicit interventions, molecular states, relations, and phenotype consequences, and they may lag behind newly published mechanistic findings. Accordingly, we start from KEGG pathways and ground benchmark instances in scientific literature to derive state-aware structured supervision and mechanistic explanations. Notably, our method is applicable to other pathway resources.

\subsection{Literature-Grounded Benchmarks for Pathway Reasoning and Relation Extraction}
Many biomedical benchmarks are derived from scientific literature, but they do not evaluate end-to-end pathway mechanism elucidation from multi-omics observations. PubMedQA focuses on evidence-based question answering over PubMed abstracts~\cite{jin2019pubmedqa}. DrugProt~\cite{luo2022sequence} and ChemProt~\cite{zhang2019chemical} evaluate local relation extraction from abstracts or sentences. However, these benchmarks do not assess biomolecular interaction inference under pathway context.

More recently, BioMaze benchmarks intervention-centric pathway reasoning and introduces pathway subgraph navigation agents such as PathSeeker~\cite{zhaobenchmarking}. However, it emphasizes navigation and subgraph operations rather than generating state-aware mechanistic chains that connect perturbed entities to pathway-level phenotypes. In contrast, BIOME-Bench provides literature-grounded, instance-level supervision with explicit molecular states, relations, and pathway phenotypes, enabling fine-grained evaluation of both interaction types and end-to-end mechanistic explanations.

\subsection{LLM-Based Systems for Multi-Omics and Pathway Interpretation}
Recent LLM-based systems such as AutoBA ~\cite{zhou2024ai} and MAPA ~\cite{ge2025leveraging} demonstrate how LLMs can support practical multi-omics workflows, including analysis planning, tool execution, and interpretation of pathway analysis outputs. However, due to the lack of end-to-end benchmarks for multi-omics pathway mechanism elucidation, model performance in this setting is often evaluated through ad hoc case studies or small manually curated datasets. This limits coverage and reproducibility, and it makes it difficult to diagnose whether models can, given only perturbed entities and pathway context, directly generate accurate pathway-mechanism explanations. To address this gap, BIOME-Bench provides literature-grounded, instance-level supervision and standardized evaluation for this end-to-end setting.

\begin{figure*}
    \centering
    \includegraphics[width=1\linewidth]{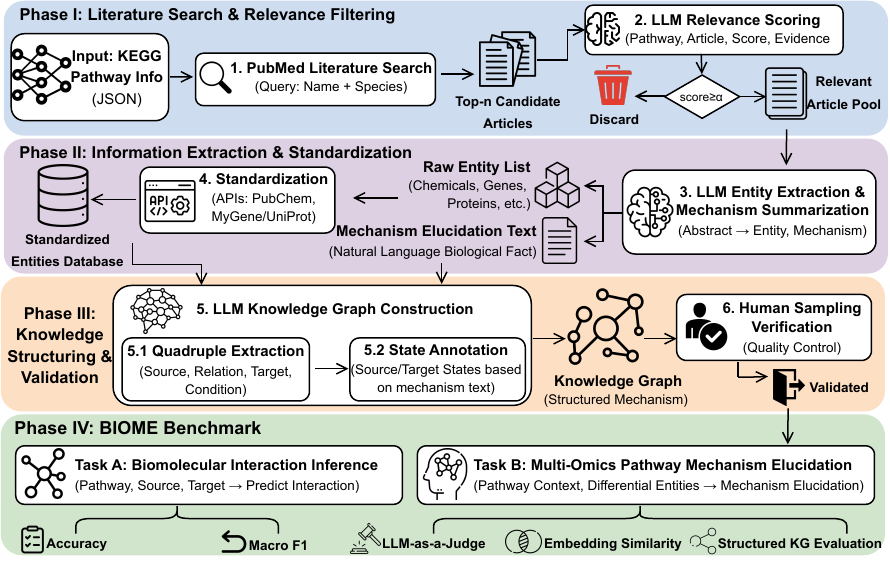}
    \caption{Workflow for constructing BIOME-Bench: (I) MeSH-guided PubMed retrieval with LLM relevance filtering; (II) LLM-based entity extraction and standardization; (III) state-aware knowledge graph construction with human sampling verification; and (IV) benchmark formulation with two tasks—biomolecular interaction inference and multi-omics pathway mechanism elucidation.}
    \label{fig:workflow}
\end{figure*}

\section{Methodology}
\label{sec:methodology}
As shown in Figure~\ref{fig:workflow}, we propose a data construction workflow that converts pathway information and supporting evidence from biomedical literature into structured, validated knowledge representations in three sequential phases (detailed prompts are provided in Appendix~\ref{app:prompts}). Building on these representations, we introduce \textbf{BIOME-Bench}, a literature-grounded benchmark for evaluating LLMs on two tasks: Biomolecular Interaction Inference and Multi-Omics Pathway Mechanism Elucidation.

\subsection{Phase I: Literature Retrieval and Relevance Filtering}
To ensure strong biological validity, the construction process begins with a rigorous literature acquisition stage. Let
\[
\mathcal{P} = \{p_1, p_2, \dots, p_n\}
\]
denote a predefined set of target KEGG pathways. Each pathway $p_i$ is characterized by its pathway name $N_{p_i}$ and associated species $S_{p_i}$.

\paragraph{MeSH-guided Literature Retrieval.}
For each pathway $p_i$, we perform a structured literature search on the PubMed ~\cite{white2020pubmed} database using Medical Subject Headings (MeSH) ~\cite{nlm:mesh-2025-browser} to improve recall precision and semantic consistency. Specifically, the pathway name $N_{p_i}$ is mapped to a set of MeSH descriptors, denoted as $\mathrm{MeSH}(N_{p_i})$, and the species $S_{p_i}$ is mapped to the corresponding MeSH organism term, denoted as $\mathrm{MeSH}(S_{p_i})$.

The final PubMed query is constructed as a conjunction of pathway-related MeSH terms and species constraints:
\begin{equation}
Q(p_i) = \mathrm{MeSH}(N_{p_i}) \;\wedge\; \mathrm{MeSH}(S_{p_i}).
\end{equation}

Executing $Q(p_i)$ yields an initial candidate document set:
\begin{equation}
D_{\text{cand}}(p_i) = \{ d_1, d_2, \dots, d_m \},
\end{equation}
where each document $d_j$ is indexed by PubMed and annotated with curated MeSH terms.

\paragraph{LLM-based Semantic and Mechanistic Relevance Scoring.}
MeSH-guided retrieval yields a controlled, high-recall candidate set, but MeSH annotations alone do not ensure that an article contains pathway-specific mechanistic evidence. To identify literature suitable for mechanism-level benchmarking, we use an LLM-based semantic evaluator with parameters $\theta$.

Given a document and pathway pair $(d, p_i)$, the evaluator assigns a relevance score
\[
s = f_{\theta}(d, p_i), \quad s \in [0,10],
\]
where $f_{\theta}$ aggregates multiple biologically motivated dimensions:
\begin{equation}
f_{\theta}(d, p_i)
= g_{\theta}\bigl(\mathbf{S}\bigr), \quad
\mathbf{S} =
\left(
\begin{array}{c}
S_{\text{subj}} \\
S_{\text{spec}} \\
S_{\text{mol}} \\
S_{\text{ctx}}
\end{array}
\right).
\end{equation}

$S_{\text{subj}}$ measures whether the pathway's biological process is the primary focus of the article, as opposed to a background mention.
$S_{\text{spec}}$ measures whether the organism studied matches the pathway species, and it also credits appropriate model organisms used to study human physiology or disease while penalizing biologically unrelated species.
$S_{\text{mol}}$ measures whether the article mentions pathway-defined molecular entities, such as genes, enzymes, or metabolites.
$S_{\text{ctx}}$ measures whether the article describes pathway regulation, such as activation, inhibition, or other modulatory effects, rather than merely reporting pathway presence.

We retain a document only if it exceeds a strict relevance threshold:
\begin{equation}
D_{\text{rel}}(p_i) =
\left\{
d \in D_{\text{cand}}(p_i)
\;\middle|\;
f_{\theta}(d, p_i) \geq \alpha
\right\}.
\end{equation}

In this work, we set $\alpha = 8$ to prioritize articles in which the target pathway is central and supported by explicit molecular and regulatory evidence.

\subsection{Phase II: Information Extraction and Entity Standardization}
\label{ssec:extraction}

\subsubsection{LLM-based Mechanistic Extraction}

For each document $d \in D_{\text{rel}}(p_i)$, we process the abstract using a LLM to produce two complementary outputs:

\begin{enumerate}
    \item \textbf{Raw Entity Set} $E_{\text{raw}}$: a collection of mentioned biological entities categorized into Chemicals, Genes/Proteins, and Phenotypes.
    \item \textbf{Mechanism Description} $M_{\text{text}}$: a coherent natural language explanation describing the molecular interactions and regulatory mechanisms reported in the document. This text later serves as ground truth for generative evaluation.
\end{enumerate}

\subsubsection{Entity Normalization and Ontology Mapping}
To ensure interoperability with external biological resources, we normalize each raw entity $e \in E_{\text{raw}}$ to a canonical identifier using an ontology resolution function $\phi(e)$. Specifically,
\begin{itemize}
    \item \textbf{Chemical entities} are mapped to PubChem~\cite{kim2016pubchem} compound identifiers (CIDs) using PubChemPy.
    \item \textbf{Genes and proteins} are mapped to NCBI Gene~\cite{brown2015gene} identifiers or UniProt~\cite{uniprot2015uniprot} accessions via MyGene.info.
\end{itemize}

To improve benchmark quality, we discard a candidate document if any entity cannot be resolved to a valid identifier. Only documents for which all entities are successfully normalized are retained, yielding the standardized entity set:
\begin{equation}
E_{\text{std}}\!=\!\{\phi(e) | e\!\in\!E_{\text{raw}} \wedge \forall e'\!\in\!E_{\text{raw}}, \phi(e')\!\neq\!\emptyset\}
\end{equation}

\subsection{Phase III: Knowledge Structuring and Validation}
This phase leverages an LLM to convert the extracted mechanistic information and standardized entities into a fine-grained, state-aware knowledge graph representation.

\subsubsection{Interaction Quadruple Extraction}

We first extract the core interaction structure from $M_{\text{text}}$. Each interaction is represented as a quadruple:
\begin{equation}
T_{\text{core}} = (e_s, r, e_t, c),
\end{equation}
where $e_s, e_t \in E_{\text{std}}$ denote the source and target entities, $r \in \mathcal{R}$ is a relation type drawn from a controlled biological vocabulary, and $c$ specifies the biological condition under which the interaction occurs.

\subsubsection{Biological State Annotation}

To capture dynamic molecular behavior, we further annotate entity-specific biological states. Let $\sigma_s$ and $\sigma_t$ denote the states of the source and target entities, respectively (e.g., mutated, overexpressed). Incorporating state information yields a state-aware hexaplet representation:
\begin{equation}
T_{\text{final}} = (e_s, \sigma_s, r, e_t, \sigma_t, c).
\end{equation}

This formulation enables the benchmark to distinguish subtle yet critical mechanistic differences, such as changes in protein abundance versus post-translational modifications.

\subsubsection{Human Expert Verification}

To establish a high-confidence gold standard, we perform human-in-the-loop validation. A randomly sampled subset of the constructed knowledge graph entries is reviewed by domain experts in molecular biology and systems biology, who cross-check each entry against the supporting literature to verify its accuracy and grounding (see Appendix~\ref{app:human_verification} for details).

\subsection{Phase IV: BIOME-Bench Task Formulation}

Based on the curated and validated knowledge representations, BIOME-Bench defines two complementary evaluation tasks.

\subsubsection{Task A: Biomolecular Interaction Inference}

This task evaluates an LLM's ability to infer precise molecular relationships within a pathway context. Given a pathway $p_i$, a source entity $e_s$ with state $\sigma_s$, a target entity $e_t$ with state $\sigma_t$, and a biological condition $c$, the model is required to predict the correct interaction relation from a finite controlled vocabulary $\mathcal{R}$:
\begin{equation}
\hat{r} = \arg\max_{r \in \mathcal{R}} P\bigl(r \mid p_i, e_s, \sigma_s, e_t, \sigma_t, c\bigr).
\end{equation}

Model performance is evaluated using Accuracy and Macro-F1 over relation labels in $\mathcal{R}$ (with invalid or unrecognized predictions treated as errors for the corresponding ground-truth label).

\subsubsection{Task B: Multi-Omics Pathway Mechanism Elucidation}

This task simulates realistic omics-driven pathway analysis scenarios. The model is provided with a pathway context $p_i$ and a set of differentially observed entities
\[
E_{\text{diff}} \subseteq E_{\text{std}},
\]
and is required to generate a coherent mechanistic explanation $\hat{Y}$ that elucidates the biological interactions, regulatory relationships, and molecular processes connecting these entities within the given pathway context.

We adopt a multi-dimensional evaluation strategy to comprehensively assess the quality of the generated explanations:
\begin{itemize}

   \item \textbf{LLM-as-a-Judge}: Given the model-generated explanation $\hat{Y}$ and the literature-derived ground truth $M_{\text{text}}$, judge model evaluates the output across four biologically motivated dimensions (scale: 1-5): Phenotype Coverage, Causal Reasoning, Factuality, and Hallucination.

    \item \textbf{Structured Knowledge Graph Evaluation}: 
        Based on the literature-derived knowledge graph, we adopt a closed-set evaluation protocol. Specifically, a extraction model is provided with the standardized knowledge graph and is only allowed to select supporting knowledge tuples from this graph based on the generated explanation $\hat{Y}$. As a result, the predicted tuple set satisfies
        \[
        \mathcal{T}_{\text{pred}} \subseteq \mathcal{T}_{\text{GT}},
        \]
        ensuring that no out-of-graph or hallucinated knowledge can be introduced.
        
        Under this constraint, factual completeness is quantified using coverage, defined as:
        \[
        \text{Coverage} = \frac{|\mathcal{T}_{\text{pred}}|}{|\mathcal{T}_{\text{GT}}|}.
        \]
    
    \item \textbf{Semantic Embedding Similarity}: We compute the cosine similarity between vector representations of $\hat{Y}$ and the ground truth mechanism text $M_{\text{text}}$ using an LLM-based embedding model, providing a complementary measure of semantic alignment.
\end{itemize}

\section{Experiments}
\label{sec:experiments}

\subsection{Benchmark Statistics and Characteristics}
\label{subsec:data_stats}

\begin{table}[h]
    \centering
    \small
    \setlength{\tabcolsep}{4.5pt}
    \renewcommand{\arraystretch}{1.15}
    \resizebox{\linewidth}{!}{
    \begin{tabular}{lrrrrr}
        \toprule
        \textbf{Species} &
        \makecell{\textbf{Number of}\\\textbf{Pathways}} &
        \makecell{\textbf{Number of}\\\textbf{Entities}} &
        \makecell{\textbf{Number of}\\\textbf{Processes and}\\\textbf{Phenotypes}} &
        \makecell{\textbf{Task A}\\\textbf{Biomolecular}\\\textbf{Interaction}\\\textbf{Inference}} &
        \makecell{\textbf{Task B}\\\textbf{Multi-Omics}\\\textbf{Pathway}\\\textbf{Mechanism}\\\textbf{Elucidation}} \\
        \midrule
        \texttt{hsa} & 80  & 1{,}349 & 1{,}781 & 4{,}032 & 490 \\
        \texttt{mmu} & 80  & 1{,}356 & 1{,}860 & 4{,}162 & 496 \\
        \texttt{rno} & 80  & 1{,}141 & 1{,}265 & 3{,}384 & 361 \\
        \midrule
        \textbf{Total} & \textbf{240} & \textbf{3{,}846} & \textbf{4{,}906} & \textbf{11{,}578} & \textbf{1{,}347} \\
        \bottomrule
    \end{tabular}
    }
    \caption{Benchmark statistics of BIOME-Bench across species.}
    \label{tab:benchmark_stats}
\end{table}

\begin{table*}[t]
\centering
\resizebox{1\textwidth}{!}{
\begin{tabular}{lccccccccc}
\toprule

\multirow{4}{*}{\textbf{Model}} & \multicolumn{2}{c}{\textbf{Biomolecular Interaction}} & \multicolumn{6}{c}{\multirow{2}{*}{\textbf{Multi-Omics Pathway Mechanism Elucidation}}} & \multirow{4}{*}{\textbf{Avg.}} \\

& \multicolumn{2}{c}{\textbf{Inference}} & \multicolumn{6}{c}{} & \\

\cmidrule(lr){2-3} \cmidrule(lr){4-9}

& \multirow{2}{*}{Acc} & \multirow{2}{*}{Macro-F1} & \multicolumn{4}{c}{LLM-as-a-Judge} & \multirow{2}{*}{Similarity} & \multirow{2}{*}{Coverage} & \\

\cmidrule(lr){4-7}

& & & Phenotype Coverage & Causal Reasoning & Factuality & Hallucination & & & \\
\midrule

Qwen3-14B       & 47.43& 43.72& 3.12& 3.31& 3.97& 4.64& 78.73& 42.38& 64.13 \\
Qwen3-32B       & 41.84& 40.51 & 3.00& 3.26& 3.89& 4.79& \textbf{78.98}& \textbf{45.43}& 63.20\\
Qwen3-235B       & 51.41& 46.21 & 3.66& 4.32& 4.54& 4.40& 77.34& 42.22& 69.45\\
DeepSeek-V3.2-R1      & 53.10& 47.52& 3.28& 4.31& 4.20& 4.10& 75.12& 40.76& 66.79\\
GLM-4.6       & 53.60& 50.08& 3.50& 4.14& 4.32& 4.18& 76.89& 39.95& 67.92\\
Gemini3-Pro       & 52.34& 46.54& 3.60& 4.57& 4.59& 4.54& 77.21& 41.13& 69.74\\
GPT-5.2          & 54.66 & \textbf{50.70} & 3.68& 4.58& 4.69& 4.62& 71.38& 37.49 & 70.70\\

Doubao-Seed-1.8       & \textbf{55.42}& 50.40& 3.81& \textbf{4.69}& 4.69& 4.57& 74.92& 39.72& 71.96\\
Intern-S1-235B        & 54.15& 50.36 & 3.96& 4.28& 4.75& \textbf{4.92}& 78.71& 44.49& 73.24\\

S1-Base-671B & 54.68 & 50.41 & \textbf{4.02}& 4.48& \textbf{4.76}& 4.83& 77.36& 44.45& \textbf{73.59}\\

\bottomrule
\end{tabular}
}
\caption{Performance comparison of large language models on BIOME-Bench. Similarity refers to the cosine similarity between the embeddings of the generated answer and the reference answer. Coverage refers to the knowledge graph coverage derived using Structured Knowledge Graph Evaluation. \textbf{Avg.} is the arithmetic mean of all metrics after normalizing each score to the $0$--$100$ range.}
\label{tab:main_results}
\end{table*}

BIOME-Bench is a multi-species benchmark that covers three commonly used organisms: \texttt{hsa} (human), \texttt{mmu} (mouse), and \texttt{rno} (rat). Table~\ref{tab:benchmark_stats} summarizes the core statistics, including the numbers of curated pathways, standardized entities, process and phenotype terms, mechanism analysis instances, and knowledge graph relations. Overall, it includes 1{,}347 instances for  multi-omics mechanism elucidation and 11{,}578 instances  for biomolecular interaction inference, both evaluated under consistent pathway contexts.

\subsection{Experimental Setup}
\label{subsec:setup}
We evaluate a range of LLMs, including Qwen3-14B, 32B and 235B~\cite{yang2025qwen3}, DeepSeek-V3.2-R1~\cite{liu2025deepseek}, GLM-4.6~\cite{glm2024chatglm}, Gemini3-Pro~\cite{google:gemini-3-pro-preview}, GPT-5.2~\cite{openai:gpt-5.2}, Doubao-Seed-1.8~\cite{seed2025seed1}, Intern-S1-235B~\cite{bai2025intern} and S1-Base-671B~\cite{ScienceOne}. We deploy Qwen3-14B, 32B and 235B with vLLM~\cite{kwon2023vllm}, and access the remaining models through their respective APIs. For all models, we conducted a single experimental run with $\texttt{temperature}=0$ and $\texttt{max\_tokens}=10240$. 

For LLM-as-a-Judge and information extraction for Structured Knowledge Graph Evaluation, we use Qwen3-32B with $\texttt{temperature}=0$ and $\texttt{max\_tokens}=10240$. For embedding-based similarity, we use Qwen3-Embedding-8B~\cite{zhang2025qwen3emb}. All experiments run on a server with $8 \times$ NVIDIA A100-SXM4-80GB GPUs.

\subsection{Main Results}
\label{subsec:main_results}
Table~\ref{tab:main_results} summarizes the overall performance of contemporary LLMs on BIOME-Bench. For biomolecular interaction inference, most models fall within a relatively narrow range (Acc $41.84\%$ to $55.42\%$, Macro-F1 $40.51\%$ to $50.70\%$), indicating that the task remains challenging and that current progress is largely incremental. Doubao-Seed-1.8 achieves the highest accuracy ($55.42\%$), whereas GPT-5.2 attains the best Macro-F1 ($50.70\%$). The discrepancy suggests that leading systems are comparable in overall correctness but differ in robustness on minority classes.

For multi-omics pathway mechanism elucidation, the LLM-as-a-Judge dimensions exhibit a consistent pattern. Factuality and hallucination control are generally strong, typically above $4$, while phenotype coverage is lower, around $3$ to $4$. This suggests that current models often produce plausible and reasonably grounded narratives, but they frequently omit required process or phenotype elements, making phenotype-level reasoning a primary bottleneck. Doubao-Seed-1.8 and GPT-5.2 score highest on causal reasoning ($4.69$ and $4.58$) while maintaining high factuality (both $4.69$). Intern-S1 performs particularly well on evidence alignment, achieving the best hallucination control ($4.92$) together with reliable phenotype coverage ($3.96$). In contrast, S1-Base-671B attains the highest phenotype coverage ($4.02$) and factuality ($4.76$), indicating an advantage in producing more complete and reliable mechanistic narratives.

In addition, similarity and knowledge graph coverage do not fully reflect judge-assessed mechanistic quality. Qwen3-32B attains the highest similarity ($0.7898$) and KG coverage ($45.43$), but it does not lead on causal reasoning or factuality. This indicates that similarity and KG coverage primarily capture proximity to reference phrasing and the breadth of entity linking, rather than the correctness and sufficiency of multi-step causal explanations. In addition, similarity can be sensitive to verbosity: stronger models often produce longer responses with auxiliary background or contextual text beyond the core mechanistic trace, which can reduce surface-form similarity despite improving explanatory content. Overall, strong BIOME performance requires balancing mechanistic completeness, causal coherence, and evidence grounding. Improvements in similarity or KG coverage therefore do not necessarily translate into better judge-assessed mechanistic reasoning.

Finally, as scientific foundation models, Intern-S1-235B and S1-Base-671B achieve stronger average performance than general-purpose models, suggesting that continued scientific-domain training has a positive effect on multi-omics pathway mechanism elucidation. Nevertheless, the two BIOME-Bench tasks remain highly challenging for current models. Existing models still struggle to reliably distinguish fine-grained relation types and to derive robust mechanistic explanations directly from perturbed entities.

\subsection{Validity of LLM-as-a-Judge}
\label{subsec:judge_validity}

\begin{figure}
    \centering
    \includegraphics[width=1\linewidth]{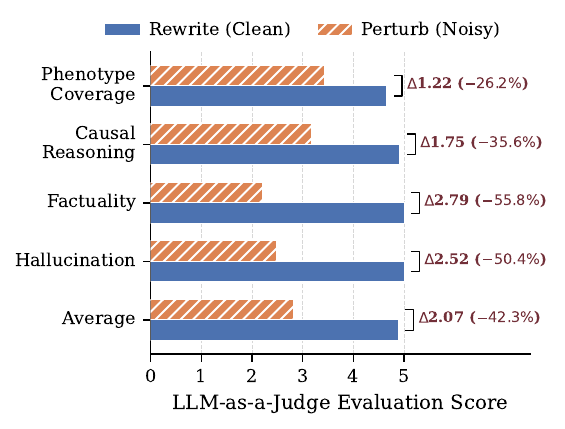}
    \caption{Sensitivity of Qwen3-32B judge to semantic perturbations. Scores are reported for rewrite and perturb . Drop\% denotes the relative score decrease from rewrite to perturb.}
    \label{fig:judge_validity}
\end{figure}

To ensure the reliability of automated evaluation in BIOME, we conducted experiments to assess the effectiveness of the LLM-as-a-judge. For each test instance, we construct two candidate answers from the gold reference. The first is a rewrite version that paraphrases the reference while preserving entities, relations, and causal semantics. The second is a perturb version that introduces targeted semantic errors by replacing key entities and/or interaction relations, while keeping the text fluent. A valid judge should remain insensitive to semantics-preserving rewrites, yet substantially penalize perturbations that degrade mechanistic correctness.

Figure~\ref{fig:judge_validity} shows that Qwen3-32B consistently distinguishes semantics-preserving rewrites from mechanistically corrupted perturbations. The rewrite answers obtain near-ceiling scores of $4.66/4.92/5.00/5.00$ on Phenotype Coverage, Causal Reasoning, Factuality, and Hallucination, with an overall average of $4.89$. In contrast, the perturb answers drop to $3.44/3.17/2.21/2.48$ and an average of $2.82$. This corresponds to relative score decreases of $26.2\%$, $35.6\%$, $55.8\%$, and $50.4\%$ for the four dimensions, and $42.3\%$ on average. The large and consistent drops indicate that the judge is highly sensitive to entity and relation perturbations even when the text remains fluent, supporting the validity of Qwen3-32B as an LLM-as-a-Judge under our reference-grounded rubric.

\subsection{Interaction Type Confusion in Biomolecular Inference}
\label{subsec:confusion_matrix}
\begin{figure}
    \centering
    \includegraphics[width=1\linewidth]{kg_relation_error_heatmap.pdf}
    \caption{Error confusion matrix for Biomolecular Interaction Inference. Rows are gold relation types and columns are predicted types. Color encodes the count of misclassified gold$\rightarrow$predicted relations.}
    \label{fig:kg_error}
\end{figure}
To analyze failure modes in \textit{Biomolecular Interaction Inference}, we aggregate all misclassified relations and summarize the confusion matrix in Figure~\ref{fig:kg_error}. Errors are dominated by coarse regulatory and causal labels. Across all mistakes, predictions most often fall into \texttt{leads\_to} (relation~$16$, 11{,}079 cases), \texttt{activates} ($1$, 10{,}469), and \texttt{inhibits} ($2$, 7{,}930), indicating a strong tendency to default to generic causality or signed regulation under uncertainty.

Two confusions account for most of this mass. First, the underspecified label \texttt{regulates} ($5$) is frequently polarized or rewritten as causation, most often to \texttt{activates} (3{,}626), \texttt{inhibits} (2{,}783), or \texttt{leads\_to} (3{,}222). Second, models blur pathway-level causation and direct regulation: gold \texttt{activates}/\texttt{inhibits} are often predicted as \texttt{leads\_to} (1{,}325/1{,}643), while gold \texttt{leads\_to} is over-interpreted as \texttt{activates} or \texttt{inhibits} (3{,}174/1{,}573). 

Overall, the results indicate that current models still fall short in distinguishing fine-grained interaction types and separating direct regulation from pathway-level causation, making it difficult to recover complete, verifiable mechanistic chains.

\section*{Limitations}

BIOME-Bench has three main limitations. First, each instance conditions on a single pathway context; extending to multi-pathway settings with explicit crosstalk and compositional reasoning is an important direction. Second, our supervision is largely organized as a one-to-one mapping between a pathway and a supporting paper; future work should build multi-pathway, multi-document mechanistic graphs that integrate dispersed evidence and enable reasoning over interconnected mechanisms. Third, our current instruction set has limited stylistic and structural diversity, while modern LLMs can be highly prompt-sensitive due to training-data distribution effects; as a result, benchmark performance may vary with prompt phrasing and introduce evaluation bias. Expanding prompt templates and reporting robustness across prompts would help mitigate this issue.

\section*{Ethics Statement}

Our benchmark construction relies on multiple public biomedical resources, including KEGG, PubMed, PubChem, NCBI Gene, and UniProt. For all resources, we strictly adhere to their respective usage policies and access them solely for academic research purposes. We also control API request rates to avoid excessive traffic and ensure responsible use. In addition, we provide explicit citations for all public biomedical resources used in this work.

Large language models are used primarily to support data processing tasks. The generated content is limited to biomedical mechanistic analysis grounded in the provided literature evidence and pathway context, and is not intended to produce offensive, hateful, or otherwise harmful content. We further incorporate validation procedures and human checks to mitigate the risk of unverified or misleading statements.

We did not employ third-party annotators. The two validators involved in expert verification are collaborators on this project, and their verification efforts constitute one component of their contribution to this work.

This study does not involve human-subject research and does not collect personal data. All information used is derived exclusively from public biomedical databases and published scientific literature.

\bibliography{main}
\appendix
\section*{Appendices}
\section{Human Expert Verification Protocol}
\label{app:human_verification}

To evaluate the reliability of our data construction workflow and to establish a high-confidence gold standard, we performed a human-in-the-loop verification study with two domain experts in molecular biology and systems biology.

\paragraph{Sampling and Materials.}
We randomly sampled 50 \textit{pathway--paper} pairs from the constructed dataset. For each pair, we provided (i) the paper abstract and the intermediate artifacts produced by our pipeline, including (ii) the LLM-extracted mechanistic text, (iii) the LLM-based relevance score with its abstract-grounded evidence and rationale, and (iv) the normalized entity list. For each normalized entity, we additionally supplied multiple database-derived aliases and synonyms to facilitate matching to the abstract. We also provided the structured outputs derived from the abstract, including annotated entity states and relations in the resulting knowledge graph.

\paragraph{Verification Steps and Criteria.}
Experts evaluated each instance using a four-step checklist. They were instructed to use the abstract as the evidence source, which matches the evidence constraints of our construction pipeline:
\begin{enumerate}[leftmargin=1.5em]
    \item \textbf{Pathway--paper relevance.} Based on the abstract, determine whether the paper is highly relevant to the specified pathway context.
    \item \textbf{Hallucination check for mechanistic text.} Determine whether the LLM-extracted mechanistic text contains unsupported claims, defined as statements not substantiated by the abstract.
    \item \textbf{Entity presence validation.} Verify whether each normalized entity is present in the abstract, allowing matches via the provided aliases and synonyms.
    \item \textbf{State and relation grounding.} Verify whether the annotated molecular states and inter-entity relations in the knowledge graph are consistent with the abstract.
\end{enumerate}

We labeled an instance as \textbf{reliable} only if it satisfied all four criteria. Otherwise, it was labeled as unreliable.

\paragraph{Results.}
All 50 sampled instances satisfied the checklist criteria (100\% pass rate). This result supports that our workflow produces robust intermediate artifacts and accurate structured supervision.

\section{Case Study: Multi-Omics Pathway Mechanism Elucidation}
\label{app:case_study}
\CaseLegend

We present a representative end-to-end instance of multi-omics pathway mechanism elucidation for the KEGG pathway \textit{Osteoclast differentiation} (hsa04380), grounded in a linked mechanistic study (PMID: 41266497). The instance is specified as follows:

\paragraph{Input.}
\textbf{Pathway Name:}
Osteoclast differentiation -- Homo sapiens (human). \\

\textbf{Pathway Description:}
The osteoclasts, multinucleared cells originating from the hematopoietic monocyte-macrophage lineage, are responsible for bone resorption. Osteoclastogenesis is mainly regulated by signaling pathways activated by RANK and immune receptors, whose ligands are expressed on the surface of osteoblasts. Signaling from RANK changes gene expression patterns through transcription factors like NFATc1 and characterizes the active osteoclast. \\

\textbf{Observed Entities:}
Gsk-650394, SGK1, STAT3 (dephosphorylated), MYCL (expression reduced), CTSK (transcription promoted), NFATC1. \\

\textbf{Instruction:}
Based on the pathway context and biological principles, generate a coherent mechanism analysis explaining how these entities interact to produce the observed phenotype.

\paragraph{Output.}
Mechanistically, \ENT{SGK1} \REL{phosphorylates} \ENT{STAT3} at Tyr705, and Tyr705-phosphorylated \ENT{STAT3} \REL{upregulates} \ENT{MYCL} expression. \ENT{MYCL} directly \REL{binds} the \ENT{CTSK} promoter and \REL{promotes} \ENT{CTSK} transcription, thereby supporting \PHE{osteoclast differentiation} and \PHE{bone resorption}. Consistent with this signaling axis, pharmacological inhibition of \ENT{SGK1} by \ENT{GSK650394} reduces \ENT{STAT3} Tyr705 phosphorylation, leading to reduced \ENT{MYCL} expression and impaired osteoclastogenesis. Importantly, \ENT{MYCL} overexpression can \REL{rescue} the osteoclast differentiation defect induced by \ENT{SGK1} inhibition, providing functional support for the causal \ENT{SGK1}$\rightarrow$\ENT{STAT3}$\rightarrow$\ENT{MYCL}$\rightarrow$\ENT{CTSK} pathway in osteoclastogenesis.

This instance specifies a chemical perturbation (\ENT{GSK650394}) and heterogeneous molecular alterations spanning signaling and transcriptional regulation, including \ENT{SGK1}, \ENT{STAT3} \STA{(dephosphorylated at Tyr705)}, \ENT{MYCL} \STA{(reduced expression)}, and \ENT{CTSK} \STA{(increased transcription)}, together with the pathway-context transcription factor \ENT{NFATC1}. It is challenging because the model must integrate these observations into a single, causally constrained explanation that links a targeted intervention to pathway-level phenotypes \PHE{osteoclast differentiation} and \PHE{bone resorption}. In particular, the model must (i) perform state-aware grounding to align entities with explicit molecular states (e.g., Tyr705 phosphorylation of \ENT{STAT3}) rather than treating them as state-free symbols, (ii) propagate intervention effects with correct directionality under \ENT{SGK1} inhibition, (iii) construct a multi-step causal chain from upstream kinase signaling to transcriptional control and phenotype rather than enumerating isolated facts, and (iv) incorporate pathway priors such as \ENT{NFATC1} as contextual constraints without asserting unsupported causal edges. Although these requirements reflect routine mechanistic reasoning in biology, contemporary models often fail to produce end-to-end explanations that are simultaneously state-consistent, intervention-consistent, and evidence-grounded from the provided perturbed entities and pathway context.

\clearpage
\section{Prompts for Data Construction Workflow}
\label{app:prompts}

\begin{tcolorbox}[
  breakable,
  enhanced,
  width=\textwidth,
  fontupper=\ttfamily\scriptsize,
  title={\small Prompt for Semantic and Mechanistic Relevance Scoring},
  listing only,
  listing options={basicstyle=\ttfamily\scriptsize,breaklines=true,columns=fullflexible}
]
\scriptsize

\textbf{\# Role}\\
You are an expert Biocurator and Molecular Biologist specializing in pathway analysis and literature mining.
Your task is to evaluate the relevance between a specific KEGG Pathway and a scientific article (PubMed).

\vspace{0.5em}
\textbf{\# Task}\\
Analyze the provided ``KEGG Pathway Info'' and ``Literature Info'' to determine if the article provides meaningful evidence, context, or experimental data related to the pathway.

\vspace{0.5em}
\textbf{\# Input Data}

\textbf{\#\# 1. KEGG Pathway Info}\\
\{\{KEGG\_PATHWAY\_JSON\}\}

\textbf{\#\# 2. Literature Info}\\
\{\{PAPER\_JSON\}\}

\vspace{0.5em}
\textbf{\# Evaluation Criteria \& Steps}
\begin{enumerate}
  \item \textbf{Subject Matching}: Does the article primarily discuss the biological process described in the pathway (e.g., Glycolysis, Gluconeogenesis)?
        Differentiate between core focus vs.\ background mention.
  \item \textbf{Species Consistency}: Check if the species in the pathway (e.g., ``hsa'' for Human) matches the organism model in the paper.
        \begin{itemize}
          \item \textit{Note}: If the pathway is Human but the paper uses a model organism (e.g., Mouse/Murine) to simulate human physiology/disease, this is considered \textbf{Relevant}.
          \item \textit{Note}: If the species are completely unrelated (e.g., Plant pathway vs.\ Human study), penalize the score.
        \end{itemize}
  \item \textbf{Molecular Evidence}: Look for specific mentions of the genes, enzymes, or metabolites described in the pathway description.
  \item \textbf{Directionality/Context}: Does the paper discuss the activation, inhibition, or regulation of this pathway?
\end{enumerate}

\vspace{0.5em}
\textbf{\# Scoring Standard (0--10)}
\begin{itemize}
  \item \textbf{0--1 (Irrelevant)}: No meaningful connection. The terms might appear only in references or unrelated contexts.
  \item \textbf{2--4 (Low)}: Pathway is mentioned as a keyword or broad concept, but not investigated. Major species mismatch without translational value.
  \item \textbf{5--7 (Medium)}: The pathway is part of the results (e.g., ``we observed changes in glycolysis''). Valid species match or relevant model.
  \item \textbf{8--10 (High)}: The pathway is the central topic. The paper elucidates mechanisms, regulation, or disease implications of this specific pathway. High species alignment.
\end{itemize}

\vspace{0.5em}
\textbf{\# Output Format}\\
Provide the result in a valid JSON object strictly adhering to the following structure.
Do not output markdown backticks (\verb|```|) or extra text.

\begin{verbatim}
{
  "relevance_score": <int, 0-10>,
  "relevance_level": "<High/Medium/Low/Irrelevant>",
  "species_check": "<Briefly state if species match or if a valid model organism is used>",
  "evidence_summary": "<Extract 1-2 key sentences from the abstract that support the link>",
  "reasoning": "<Concise explanation of the score, focusing on biological mechanisms and study focus>"
}
\end{verbatim}

\end{tcolorbox}

\clearpage
\begin{tcolorbox}[
  breakable,
  enhanced,
  width=\textwidth,
  fontupper=\ttfamily\scriptsize,
  title={\small Prompt for Entity Extraction and Mechanism Summarization},
  listing only,
  listing options={basicstyle=\ttfamily\scriptsize,breaklines=true,columns=fullflexible}
]
\scriptsize

\textbf{\# Role}\\
You are an expert Systems Biologist and Biomedical Literature Curator.
Your task is to extract, categorize, and standardize key molecular information to construct a high-quality ``Ground Truth'' benchmark for multi-omics pathway analysis.

\vspace{0.5em}
\textbf{\# Goal}\\
You will be provided with a \textbf{Target Pathway} and a scientific paper's details.
Your goal is to generate two structured outputs:
\begin{enumerate}
  \item \textbf{Significant Entities (Classified \& Normalized)}:
  A structured list of metabolites, genes, proteins, or phenotypes explicitly mentioned in the text.
  You must classify them by type and normalize them (expand abbreviations) to facilitate downstream database mapping.
  \item \textbf{Mechanism Analysis}:
  A concise, coherent biological explanation of how these entities interact.
  Crucially, this must be written as a direct fact or expert interpretation, not as a summary of a study.
\end{enumerate}

\vspace{0.5em}
\textbf{\# Instructions}
\begin{itemize}
  \item \textbf{Entity Classification \& Normalization}
  \begin{itemize}
    \item \textbf{Chemicals/Metabolites}: Target for PubChem mapping.
    Provide the \texttt{original} text found in the abstract and a \texttt{standard\_name}
    (full chemical name, expand abbreviations like \texttt{5-ALA} to \texttt{5-aminolevulinic acid}).
    \item \textbf{Genes/Proteins}: Target for UniProt/NCBI mapping.
    Provide the \texttt{original} text and a \texttt{standard\_name} (official symbol or full protein name).
    \item \textbf{Processes/Phenotypes}: Target for GO/MeSH.
    Biological outcomes or processes (e.g., \texttt{Ferroptosis}, \texttt{Oxidative Stress}).
    Keep these separate as they are not valid targets for chemical/gene databases.
  \end{itemize}

  \item \textbf{Tone \& Style (Critical)}:
  Write the \texttt{mechanism\_analysis} as an objective biological fact.
  Do not use phrases such as ``This study reveals'', ``The authors found'', or ``We observed''.
  Start directly with the biological subject (e.g., ``Elevated levels of X cause \ldots'').

  \item \textbf{Contextualize}:
  Use the Pathway Description and Evidence Summary to filter for relevance.

  \item \textbf{Precision}:
  Extract only entities explicitly mentioned in the text.
\end{itemize}

\vspace{0.5em}
\textbf{\# Constraints}
\begin{enumerate}
  \item \textbf{No Hallucinations}:
  Do not invent standard names if an abbreviation is ambiguous.
  Use the most likely biological expansion based on context.
  \item \textbf{Relevance}:
  Focus on the \textbf{Target Pathway}.
  \item \textbf{No Meta-Language}:
  Strictly ban words referring to the source material in the mechanism analysis.
  \item \textbf{Format}:
  Return strict JSON only.
\end{enumerate}

\vspace{0.5em}
\textbf{\# Input}\\
\texttt{Target Pathway: \{\{pathway\_name\}\}}\\
\texttt{Pathway Description: \{\{pathway\_description\}\}}\\
\texttt{Title: \{\{title\}\}}\\
\texttt{Abstract: \{\{abstract\}\}}\\
\texttt{Evidence Summary: \{\{evidence\_summary\}\}}

\vspace{0.5em}
\textbf{\# Output JSON Format}\\
Do not output markdown backticks (\verb|```|) or extra text.
\begin{verbatim}
{
  "significant_entities": {
    "chemicals": [
      { "original": "Text in abstract", "standard_name": "Full Standardized Name" }
    ],
    "genes_proteins": [
      { "original": "Text in abstract", "standard_name": "Official Symbol/Name" }
    ],
    "processes_phenotypes": [
      "Phenotype 1",
      "Phenotype 2"
    ]
  },
  "mechanism_analysis": "A direct biological explanation describing the interaction of these entities and the resulting outcome."
}
\end{verbatim}

\end{tcolorbox}

\clearpage
\begin{tcolorbox}[
  breakable,
  enhanced,
  width=\textwidth,
  fontupper=\ttfamily\scriptsize,
  title={\small Prompt for Interaction Quadruple Extraction},
  listing only,
  listing options={basicstyle=\ttfamily\scriptsize,breaklines=true,columns=fullflexible}
]
\scriptsize

\textbf{\# Role}\\
You are an expert Biological Knowledge Graph Constructor.
Your task is to convert a biological mechanism text into a structured list of interactions (quadruplets) using a strict vocabulary.

\vspace{0.5em}
\textbf{\# Task}\\
Extract all biological interactions from the ``Mechanism Text'' and map them to the following JSON structure:
\texttt{\{"source": "Entity A", "relation": "Relation\_Type", "target": "Entity B", "condition": "Context/Prerequisite"\}}.

\vspace{0.5em}
\textbf{\# Constraints}
\begin{enumerate}
  \item \textbf{Entity Mapping}:
  Use the exact names provided in the ``Standardized Entities List'' for \texttt{source} and \texttt{target}.
  \begin{itemize}
    \item Note the \textbf{[Type]} tags provided in the input (e.g., \texttt{[Chemical]}, \texttt{[Gene]}, \texttt{[Phenotype]}) to understand the biological context of each entity.
    \item If an entity is missing from the list but critical for the logic, use its name from the text.
  \end{itemize}

  \item \textbf{Relation Vocabulary (Strict)}:
  Use only the specific relation strings defined in the lists within the following JSON schema.
  Use the inline comments (\texttt{//}) as guidance.

\begin{verbatim}
{
  "Regulatory": [
    "activates",                 // covers: activation
    "inhibits",                  // covers: inhibition
    "upregulates_expression",    // covers: expression
    "downregulates_expression",  // covers: repression
    "regulates"                  // covers: indirect effect, state change
  ],
  "Physical_Interaction": [
    "binds",                     // covers: binding/association
    "dissociates_from"           // covers: dissociation
  ],
  "Modification": [
    "phosphorylates",            // covers: phosphorylation
    "dephosphorylates",          // covers: dephosphorylation
    "ubiquitinates",             // covers: ubiquitination
    "glycosylates",              // covers: glycosylation
    "methylates"                 // covers: methylation
  ],
  "Metabolic": [
    "produces",                  // covers: compound (enzyme -> product)
    "consumes",                  // covers: metabolic_reaction (substrate -> enzyme)
    "converts_to"                // covers: metabolic_reaction (substrate -> product)
  ],
  "Causal (Phenotypic)": [
    "leads_to",                  // e.g., leads to Apoptosis
    "increases_level",           // e.g., increases ROS levels
    "decreases_level"
  ]
}
\end{verbatim}

  \item \textbf{Compound Handling}:
  If ``Gene A increases Metabolite B to activate Gene C'', split it into:
  \begin{itemize}
    \item \texttt{(Gene A, increases\_level, Metabolite B)}
    \item \texttt{(Metabolite B, activates, Gene C)}
  \end{itemize}
  Do not use \texttt{compound} as a relation.

  \item \textbf{Condition Extraction}:
  Extract a short, specific phrase describing when, where, or how the interaction happens (e.g., ``upon light activation'',
  ``in HO-1 deficient cells'', ``under hypoxia'', ``when combined with DCA'').
  If the interaction is a general biological fact with no specific constraint, use \texttt{"General"}.
\end{enumerate}

\vspace{0.5em}
\textbf{\# Input}\\
\texttt{Mechanism Text: \{\{mechanism\_text\}\}} \\
\texttt{Standardized Entities List: \{\{formatted\_entity\_str\}\}}

\vspace{0.5em}
\textbf{\# Output Format}\\
Return a strict JSON object following this exact schema:

\begin{verbatim}
[
  {
    "source": "Entity A",
    "relation": "Relation_Type_From_Vocabulary",
    "target": "Entity B",
    "condition": "Context string or 'General'"
  }
]
\end{verbatim}

\end{tcolorbox}

\clearpage
\begin{tcolorbox}[
  breakable,
  enhanced,
  width=\textwidth,
  fontupper=\ttfamily\footnotesize,
  title={\small Prompt for Biological State Annotation},
  listing only,
  listing options={basicstyle=\ttfamily\scriptsize,breaklines=true,columns=fullflexible}
]
\scriptsize

\textbf{\# Role}\\
You are an expert Biological Graph Annotator.

\vspace{0.5em}
\textbf{\# Task}\\
You will be provided with:
\begin{enumerate}
  \item \textbf{Mechanism Text}: A biological description.
  \item \textbf{Existing Interactions}: A list of structured interactions (quadruplets) extracted from the text.
\end{enumerate}
Your job is to annotate the biological state (e.g., increased, mutated, added, phosphorylated) for the \texttt{source} and \texttt{target} entities in each interaction, based strictly on the text.

\vspace{0.5em}
\textbf{\# Instructions}
\begin{enumerate}
  \item For each interaction in the list, add two new fields: \texttt{source\_state} and \texttt{target\_state}.
  \item Extract the state as a short phrase from the text.
  \begin{itemize}
    \item Examples: \texttt{"elevated levels"}, \texttt{"administration"}, \texttt{"mutated"}, \texttt{"deficiency"}, \texttt{"accumulation"}, \texttt{"overexpression"}.
  \end{itemize}
  \item \textbf{Treatment/Drug Handling}:
  If an entity is a drug or treatment added to the system, use terms such as \texttt{"administration"}, \texttt{"added"}, or \texttt{"treated with"}.
  \item \textbf{Default}:
  If the text does not specify a change or state for that entity in that specific context (i.e., it is simply a pathway component), use \texttt{"Present"} or \texttt{"Endogenous"}.
  \item \textbf{Constraint}:
  Do not change the original \texttt{source}, \texttt{relation}, \texttt{target}, or \texttt{condition} values. Keep them exactly as provided.
\end{enumerate}

\vspace{0.5em}
\textbf{\# Input}\\
\texttt{Mechanism Text: \{\{mechanism\_text\}\}} \\
\texttt{Existing Interactions to Annotate: \{\{existing\_graph\_json\}\}}

\vspace{0.5em}
\textbf{\# Output Format}\\
Return a strict JSON object with a single key \texttt{"annotated\_kg"} containing the updated list.

\vspace{0.5em}
\textbf{\# Example Output Schema}
\begin{verbatim}
{
  "annotated_kg": [
    {
      "source": "Entity A",
      "source_state": "elevated levels",
      "relation": "increases_level",
      "target": "Entity B",
      "target_state": "accumulation",
      "condition": "General"
    }
  ]
}
\end{verbatim}

\end{tcolorbox}

\end{document}